\documentclass[10pt,twocolumn,letterpaper]{article}

\usepackage{wacv}

\usepackage{graphicx}
\usepackage{amsmath}
\usepackage{amssymb}
\usepackage{booktabs}
\usepackage{graphicx}
\usepackage{caption}
\usepackage[export]{adjustbox}
\usepackage{float} 
\usepackage[table]{xcolor}  

\usepackage[pagebackref,breaklinks,colorlinks]{hyperref}
\usepackage{array}

\usepackage[capitalize]{cleveref}
\crefname{section}{Sec.}{Secs.}
\Crefname{section}{Section}{Sections}
\Crefname{table}{Table}{Tables}
\crefname{table}{Tab.}{Tabs.}

\def\wacvPaperID{2335} 
\def\confName{WACV}
\def\confYear{2025}

\begin{document}

\title{Joint Audio-Visual Idling Vehicle Detection \\ with Streamlined Input Dependencies}


\author{%
\begin{tabular}{@{}c@{}}
Xiwen Li$^{\star}$ \qquad 
Rehman Mohammed$^{\star}$ \qquad 
Tristalee Mangin$^{\dagger}$ \qquad 
Surojit Saha$^{\star}$\\ 
Kerry Kelly$^{\dagger}$ \qquad 
Ross Whitaker$^{\star}$ \qquad 
Tolga Tasdizen$^{\star}$ \\
\\
$^{\star}$ Scientific Computing and Imaging Institute, University of Utah, United States \\
$^{\dagger}$ Department of Chemical Engineering, University of Utah, United States
\end{tabular}
}



\maketitle

\begin{abstract}
Idling vehicle detection (IVD) can be helpful in monitoring and reducing unnecessary idling and can be integrated into real-time systems to address the resulting pollution and harmful products. The previous approach\cite{Li2023RealTimeIV}, a non-end-to-end model, requires extra user clicks to specify a part of the input, making system deployment more error-prone or even not feasible. In contrast, we introduce an end-to-end joint audio-visual IVD task designed to detect vehicles visually under three states: moving, idling and engine off. Unlike feature co-occurrence task such as audio-visual vehicle tracking, our IVD task addresses complementary features, where labels cannot be determined by a single modality alone. To this end, we propose AVIVDNet, a novel network that integrates audio and visual features through a bidirectional attention mechanism. AVIVDNet streamlines the input process by learning a joint feature space, reducing the deployment complexity of previous methods. Additionally, we introduce the AVIVD dataset, which is seven times larger than previous datasets, offering significantly more annotated samples to study the IVD problem. Our model achieves performance comparable to prior approaches, making it suitable for automated deployment. Furthermore, by evaluating AVIVDNet on the feature co-occurrence public dataset MAVD \cite{MM-Distill}, we demonstrate its potential for extension to self-driving vehicle video-camera setups.

\end{abstract}

\section{Introduction}
\label{sec:intro}

\begin{figure}[h!]
    \centering
    \includegraphics[width=\linewidth]{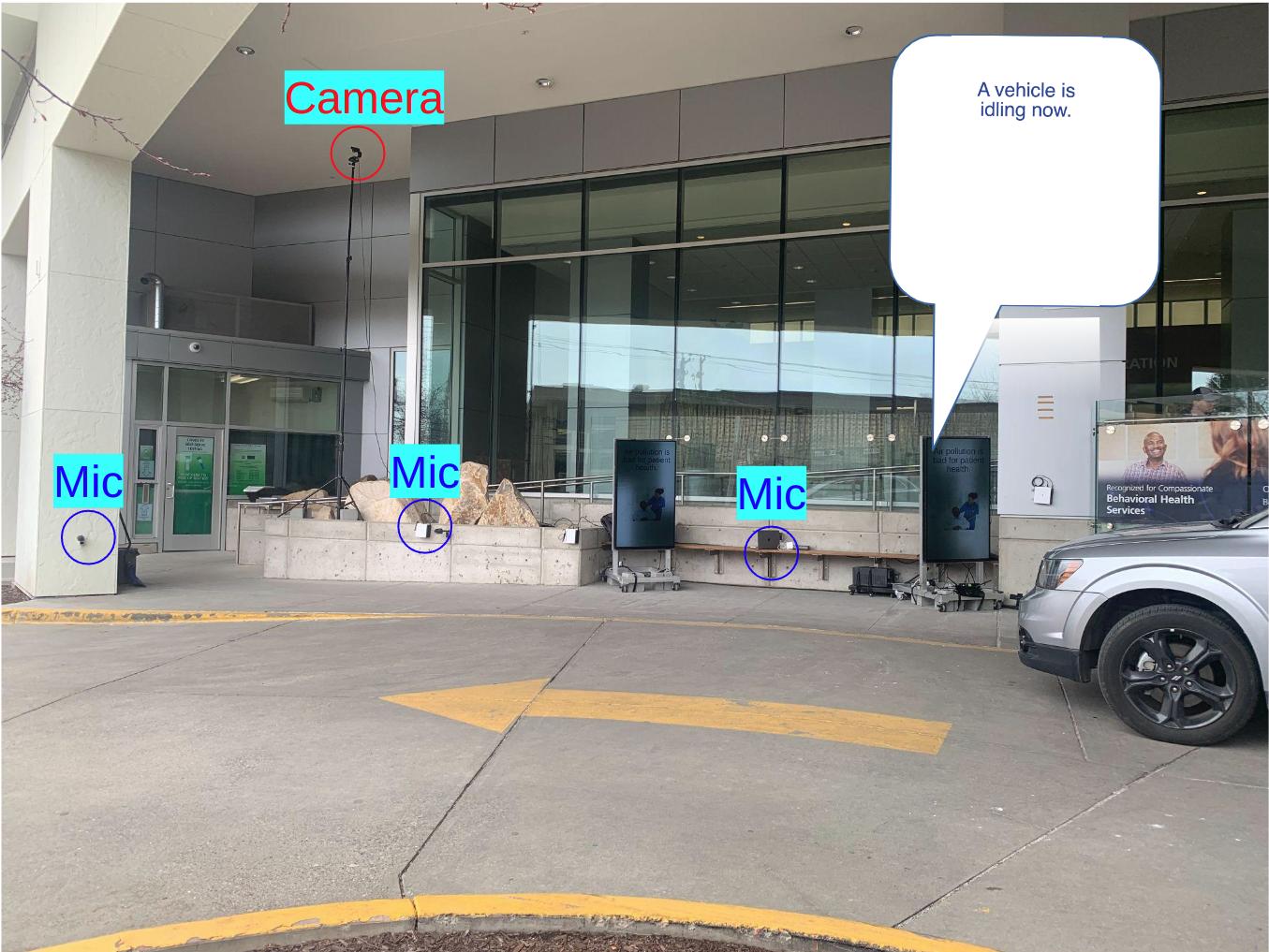} 
    \caption{Experimental Setup. We positioned 6 microphones along the roadside and installed a webcam approximately 20 feet above the ground (3 of them are not shown in the picture). An ITS detecting idling vehicles displays a reminder message on the screen.}
    \label{fig:setup}
\end{figure}

An idling vehicle is defined as one with its engine running while stationary, often occurring during parking or waiting, such as in pick-up or drop-off scenarios. As part of an Intelligent Transportation System (ITS), idling vehicle detection (IVD) plays a key role in addressing anti-idling events. An ITS that detects idling vehicles in areas like gas station queues or hospital pick-up zones could help create a dynamic feedback system to remind drivers to minimize excessive idling, as shown in \cref{fig:setup}. Reducing unnecessary idling behavior offers several benefits: (1) mitigating poor air quality that negatively impacts public health, (2) reducing greenhouse gas emissions that contribute to global warming, and (3) lowering fuel consumption and engine wear. To develop a public ignition status monitoring system, remote sensing technologies like infrared (IR) cameras have been proposed for temporal object detection tasks \cite{Bastan2018RemoteDO}. However, IR cameras are expensive and difficult to integrate with standard computing systems as it requires specialized hardware, making the deployment of such ITS solutions less economical and practical. As an alternative, using portable microphones (audio) and a web camera \cite{Li2023RealTimeIV} (video) provides a more efficient and accessible solution for visual IVD, effectively reframing it as an audio-visual detection problem.

In current audio-visual learning tasks, audio and visual features typically coexist in the same domain. Common problems like active speaker detection (ASD) \cite{Roth2019AvaAS} using single-channel movie audio, or outdoor vehicle tracking \cite{MM-Distill} with a multichannel microphone array, have been addressed through fusion networks employing convolutional neural networks, graph networks, or transformers. These tasks assume that audio and visual features are either both present or both absent in any given instance within the dataset. In other words, there is a strong correlation between the presence of features across both modalities. For example, an active speaker in a video will exhibit both lip movement and voice, while a non-active speaker will have neither. Similarly, vehicle tracking combines visual appearance or motion with the sound of the engine. Another research direction leverages this feature co-occurrence, where a video-based teacher model is used to train an audio-only vehicle detector for operation in low-light conditions. The mainstream approach for solving multimodal problems involves using encoders for both audio and video inputs, followed by feature fusion. However, we observe that architectures focusing on feature complementary problems have not been thoroughly explored. In previously proposed IVD research \cite{Li2023RealTimeIV}, idling or engine-off vehicles represent a feature complementary challenge, where a single modality is insufficient to determine the vehicle’s status. Instead of using a feature-combining network, their approach requires video, audio, and user-provided visual coordinates of microphones to predict bounding boxes and labels. The method first employs a 3D CNN to detect whether a vehicle is moving or static. For static vehicles, it heuristically identifies the closest microphone to each vehicle using bounding box coordinates and pre-specified microphone pixel locations from the user input. Then, each microphone channel is classified as either having or not having engine sound, and the results are fused to generate the final labels. This late fusion approach introduces the need for additional user input to specify the nearest microphone on the image, which can be error-prone when deployed. This raises the research question: can we develop an end-to-end model that integrates audio and video features with streamlined user inputs? Such a model would need to learn the correlations between modalities in feature complementary scenarios.

We approach this problem in two steps. First, we introduce the feature complementary task of IVD, which classifies each vehicle as moving, idling, or engine off, following the method in \cite{Li2023RealTimeIV}. In this new detection problem, each vehicle is not only detected but also categorized into one of three distinct classes based on specific feature correspondences: (1) moving: characterized by both motion and sound, (2) idling: stationary with sound, and (3) engine off: stationary without any car sound. The unique aspect of this problem is that a static vehicle in the video may or may not have corresponding features in the audio domain, making it fundamentally different from traditional vehicle tracking tasks. As a result, it cannot be effectively solved using a single modality network. Second, to address this challenge, our proposed method focuses on learning and aligning features across modalities by constructing a joint feature space. This is achieved through the use of reconstruction and attention modules to capture correlations in the spatial dimension.

In this paper, building on the setup introduced in \cite{Li2023RealTimeIV}, we make three main contributions: 
\begin{itemize}
    \item We introduce a large-scale AVIVD dataset specifically designed for IVD.
    \item We propose a novel joint audio-visual model, AVIVDNet, which streamlines input requirements for more efficient processing.
    \item Our proposed model demonstrates performance comparable to state-of-the-art methods on both the AVIVD and MAVD datasets.
\end{itemize}

\begin{figure*}[h!]
    \centering
    \includegraphics[width=\textwidth]{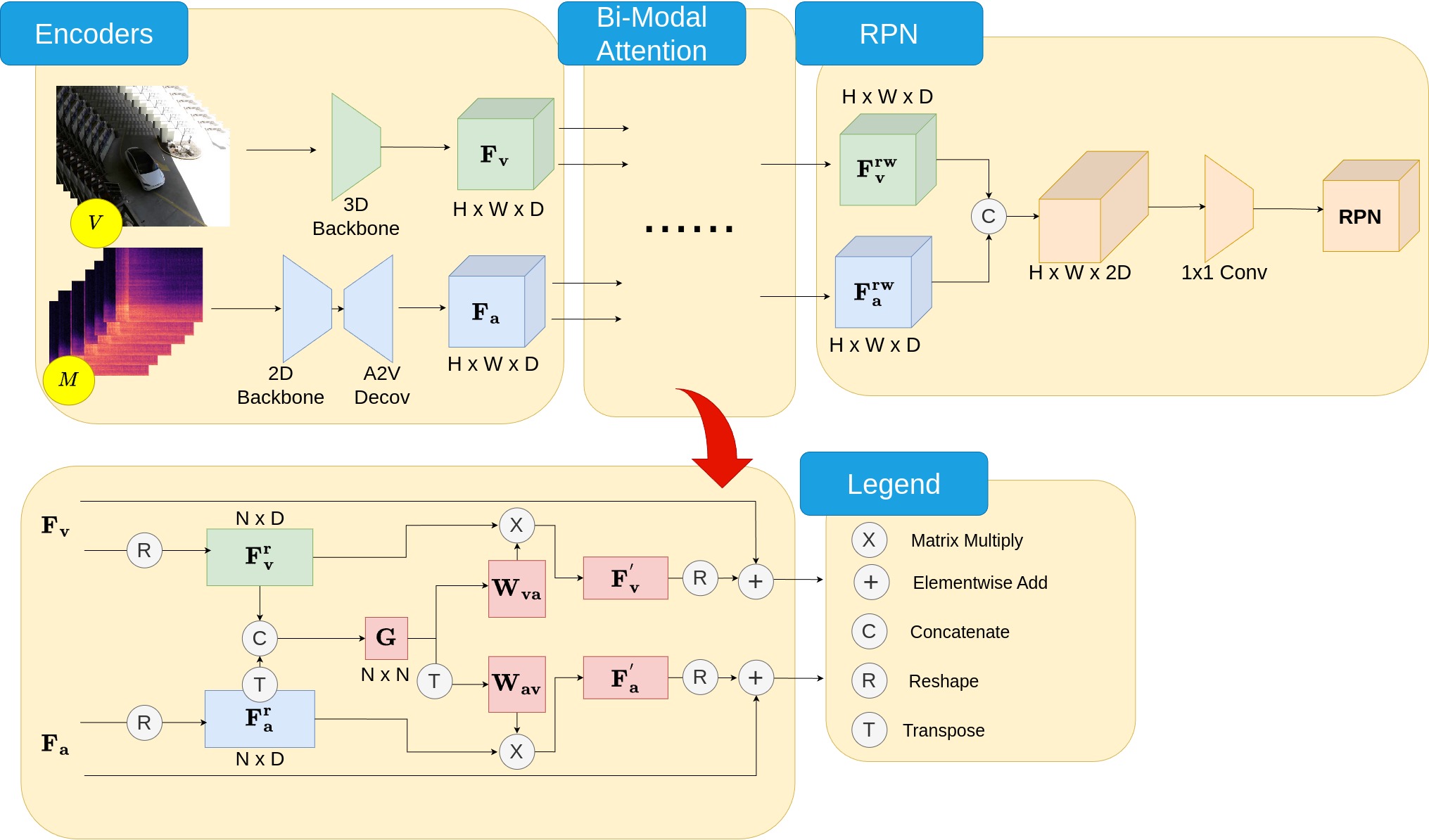} 
    \caption{Algorithm workflow. Our algorithm consists of an encoding module, a bidirectional attention module, and a region proposal network.}
    \label{fig:algorithm}
\end{figure*}

\section{Related Work}
Audio-visual learning is a rapidly growing area in computer vision, focused on how the integration of audio and visual information can enhance performance across various tasks. By leveraging both modalities, models gain richer contextual information. Common applications include segmentation \cite{zhou2022avs, Pan2022WnetAV}, localization \cite{geng2023dense}, action recognition \cite{Lee2021CrossAttentionalAF}, and representation learning \cite{Lin2022VisionTA}. A subset of audio-visual learning focuses specifically on detection tasks. For example, methods like \cite{afouras2021selfsupervised, tian2018ave} target general object detection, while others, such as \cite{Liao2023ALW, Alcazar2022EndtoEndAS, Min2022LearningLS, Zhang2022UniConIS, Jung2023TalkNCEIA}, address active speaker detection (ASD) \cite{Roth2019AvaAS}. ASD models use sequences of cropped faces from movie clips, along with synchronized audio, to predict whether a person is speaking in each frame. Another group of works focuses on vehicle detection, such as \cite{Gan2019SelfSupervisedMV, Zrn2022SelfSupervisedMV, TowardsRobustAudioBasedVehicleDetectio, MM-Distill, Zrn2022SelfSupervisedMV}. These methods often rely on knowledge distillation pipelines, where an audio-based student detector is trained using a visual-based teacher model. This approach allows audio detectors to learn from visual data, enabling them to function in low-light conditions where visual detectors struggle, such as at night. These outdoor tasks typically depend on phase differences from multichannel microphones to help localize vehicles. Among these, \cite{MM-Distill} introduced the MAVD dataset for knowledge-distillation research. However, tasks like ASD and vehicle tracking are highly dependent on the co-existence of features in both modalities. The idling vehicle detection (IVD) problem presents a different challenge, as an engine-off vehicle is visually stationary but lacks any audio signature. A single audio detector cannot accurately detect non-idling vehicles. Therefore, IVD models must incorporate features from both vehicle motion and the presence (or absence) of engine sound. Consequently, our model does not rely on knowledge distillation and instead focuses on jointly integrating these complementary features for a more comprehensive detection approach.

\subsection{Idilng Vehicle Detection} 
Visual idling vehicle detection (IVD) was recently introduced by \cite{Bastan2018RemoteDO} and \cite{Li2023RealTimeIV}. \cite{Bastan2018RemoteDO} proposes an infrared imaging-based Faster R-CNN to detect the heat generated by the engine block, relying on changes in the engine's heat signature. Alternatively, \cite{Li2023RealTimeIV} introduces a 3D-CNN for visual motion detection and a 2D-CNN for classifying audio engine sounds to detect idling vehicles. While the IR-imaging approach is straightforward, it presents several significant drawbacks: (1) High latency, as heat accumulation and dissipation take time, and many infrared cameras operate at low frame rates (e.g., capturing one image every five seconds). (2) Initial experiments in \cite{Li2023RealTimeIV} show that direct sunlight or high ambient temperatures can interfere with the model's ability to detect a hot engine block. (3) As noted by the authors \cite{Bastan2018RemoteDO}, the engine block and exhaust pipe are primary heat sources; however, false positives and negatives occur when the engine block is not in the camera's field of view, leading to unreliable detection. Preliminary results also suggest inconclusive outcomes when detecting heat from vehicle exhaust (rear of the vehicle). (4) Infrared cameras are generally expensive and challenging to deploy in diverse environments. Moreover, the process of detecting changes in heatmaps is slow and can be easily affected by surrounding heat sources, making it unsuitable for timely detection. Timeliness is a critical factor in idling vehicle detection, so we build upon the camera and microphone setup from \cite{Li2023RealTimeIV}, which avoids the complexities of IR-based detection. However, the approach in \cite{Li2023RealTimeIV} requires manual user input to specify microphone coordinates on the image, which introduces the risk of errors and complicates deployment. To address this, our method eliminates the need for user interaction, creating a fully automated system that improves the reliability and ease of deployment.


\section{Proposed Method}

\subsection{Problem Definition}
Following \cite{Li2023RealTimeIV}, we define the idling vehicle detection (IVD) problem as localizing and classifying the status of each vehicle in the final frame of a video clip $V$. Classes are $Y \in {Y_{\rm moving}, Y_{\rm idling}, Y_{\rm eoff}}$, given 6-channel audio input $M$ and $V$. Here, $Y_{\rm moving}$ indicates the vehicle is in motion, $Y_{\rm idling}$ means the vehicle is stationary with the engine running, and $Y_{\rm eoff}$ represents the vehicle is stationary with the engine off.

When a vehicle is moving, it exhibits both motion and sound. In the idling state, the vehicle has no motion but produces sound. When the engine is off, the vehicle lacks both motion and sound. In this problem, visual and audio features are complementary, as a single modality alone is insufficient for reliable detection. Audio alone, for instance, cannot distinguish all vehicle states accurately. The previous approach used the video clip $V$, audio input $M$, and a microphone visual dictionary $L$ to estimate vehicle motion, match the closest microphone channel, classify engine status, and fuse the final predictions.

In contrast, we propose a novel and unified network, AVIVDNet, which only requires $V$ and $M$ as inputs. This end-to-end algorithm consists of three modules, as demonstrated in \cref{fig:algorithm}.

\subsection{Audio to Visual Mapping Network}
In the feature encoding stage, the inputs $V$ (video) and $A$ (audio) are processed by a 3D CNN and a 2D CNN, respectively. The 3D CNN captures the motion status of each vehicle in the final frame, producing a feature map $\mathbf{F_{v}}$, where each spatial element corresponds to a specific region of the last image, encoding vehicle object features similar to the approach used in \cite{kopuklu2019yowo}. We observed that using complex encoders, such as ResNeXt-101, led to overfitting on our dataset due to their high complexity. As a result, we opted for the lightweight 3D MobileNet architecture \cite{Kpkl2019ResourceE3} for motion encoding. For audio embedding, we employ the same pretrained 2D MobileNet \cite{howard2019searching}. From the audio embedding, we apply deconvolution layers to obtain the feature map $\mathbf{F_{a}}$. Previous work \cite{Li2023RealTimeIV} highlighted the difficulty of encoding vehicle engine sounds due to the wide variety of real-world engine noises and the limited data available for training. To address this, they employed a contrastive ResNet50 latent space pretrained on a public audio dataset. In our experiments, we explored a similar approach using ResNet50 for audio feature extraction.

A straightforward approach to learning the relationship between microphone channels and image spatial objects is to leverage cross-attention. However, since attention modules typically require large amounts of data to train effectively, we found that cross-attention performed poorly due to the limited size of our dataset. Inspired by the audio detection network in \cite{TowardsRobustAudioBasedVehicleDetectio}, we propose a network that maps audio features directly to image features in the spatial domain. In this pipeline, each sounding vehicle is mapped from the mel-spectrogram’s power and semantic feature distribution to a spatial audio feature map $ \mathbf{F_{a}}$, which shares the same spatial dimensions as the visual feature map $\mathbf{F_{v}}$. In this spatial representation, vehicles with the engine off are characterized by silence or environmental sound embeddings, while idling vehicles exhibit engine sound features in the audio spatial space.

\subsection{Bidirectional Audio Visual Attention}
One potential solution to replacing the heuristic search in \cite{Li2023RealTimeIV} is the use of transformer cross-attention or a full transformer architecture. However, we found that training with heavy multihead attention blocks led to encoder overfitting due to the limited size of our dataset. To address this, we propose a bidirectional cross-modality attention module, inspired by \cite{Fu2018DualAN}, to associate the visual feature map with the audio spatial map. After the audio and visual features are extracted by the encoders, as represented by \cref{eq:reshape}, the video feature $\mathbf{F_{v}}$ and the audio feature $\mathbf{F_{a}}$ are flattened along their spatial dimensions for further processing.

\begin{equation}
\label{eq:reshape}
\begin{aligned}
        \mathbf{F_{v}}\in \mathcal{R}^{H\times W \times D}\xrightarrow{reshape} \mathbf{F_{v}^{r}}\in \mathcal{R}^{N\times D} \\
         \mathbf{F_{a}}\in \mathcal{R}^{H\times W \times D}\xrightarrow{reshape}  \mathbf{F_{a}^{r}}\in \mathcal{R}^{N\times D}
    \end{aligned}
\end{equation}
Then we compute the gram matrix acros two modalities \cref{eq:gram}, where dot products of each row are computed. In this way correlations between two feature vectors can be calculated.
\begin{equation}
\label{eq:gram}
    \mathbf{G}\in \mathcal{R}^{N\times N} =\mathbf{F_{v}^{r}} \cdot (\mathbf{F_{a}^{r}})^{T}
\end{equation}
Then, attention scores are computed along two directions, namely two modalities:
\begin{equation}
    \begin{aligned}
    \mathbf{W_{av}}\in \mathcal{R}^{N\times N}, \text{ where } W_{av}=\frac{exp(G_{ij})}{\Sigma_{j}^{N}exp(G_{ij})} \\
    \mathbf{W_{va}}\in \mathcal{R}^{N\times N}, \text{ where } W_{va}=\frac{exp(G_{ij}^{T})}{\Sigma_{j}^{N}exp(G_{ij})}
    \end{aligned}
\end{equation}
In each weight matrix, it computes which spatial sound and vehicle are associated with each other.
\begin{equation}
    \begin{aligned}
    \mathbf{F_{a}^{'}}=\mathbf{W_{av}}\cdot \mathbf{F_{a}^{r}} \\
    \mathbf{F_{v}^{'}}=\mathbf{W_{va}}\cdot \mathbf{F_{v}^{r}}
    \end{aligned}
\end{equation}

After both features are reweighted, they are reshaped back to spatial dimension.
\begin{equation}
    \begin{aligned}
    \mathbf{F_{a}^{'}}\xrightarrow{reshape} \mathbf{F_{a}^{''}} \\
    \mathbf{F_{v}^{'}}\xrightarrow{reshape} \mathbf{F_{v}^{''}}
    \end{aligned}
\end{equation}
After reweighed features are reshaped, each output channel $\mathbf{F_{*}^{rw}}$ is multipled with a weight factor $\gamma$ and element-wirse summed with the original feature $\mathbf{F_{*}}$.
\begin{equation}
    \begin{aligned}
    \mathbf{F_{a}^{rw}}=\gamma \cdot \mathbf{F_{a}^{''}} + \mathbf{F_{a}}\\
    \mathbf{F_{v}^{rw}}=\gamma\cdot \mathbf{F_{v}^{''}} + 
    \mathbf{F_{v}}\end{aligned}
\end{equation}

Finally, reweighted features $\mathbf{F_{a}^{rw}}$ and $\mathbf{F_{v}^{rw}}$ are concatenated along feature dimension for detection. 

\subsection{Region Proposal Network (RPN)}
The RPN predicts bounding boxes and labels for each local area. After concatenating $\mathbf{F_{a}^{rw}}$ and $\mathbf{F_{v}^{rw}}$, we apply a single $1 \times 1$ convolution layer to compute the RPN, following the same approach as \cite{kopuklu2019yowo}. The concatenated features are processed through this $1 \times 1$ convolution kernel, producing the final RPN output. The resulting output has dimensions of $\texttt{num\_anchors} \times (4\, (\texttt{bbox\_parameters})+ 1\, (\texttt{confidence})+\texttt{num\_classes})$. We calculate five anchors across all training bounding boxes using KMeans, as outlined in \cite{kopuklu2019yowo}.


\subsection{Loss Function}
Following \cite{kopuklu2019yowo, Redmon2016YOLO9000BF}, our loss function is composed of focal loss for classification, smooth L1 loss for bounding box regression, and mean squared error (MSE) loss for confidence regression.
$$l_{total}=l_{Focal}+l_{x}+l_{y}+l_{w}+l_{h}+l_{conf}$$

\section{Experiments}

\subsection{Datasets}
Our experiments are conducted on both the AVIVD and MAVD datasets \cite{MM-Distill}. Although both datasets are vehicle-related, they represent two distinct setups: AVIVD is based on a surveillance camera system (\cref{fig:setup}), while MAVD uses an in-vehicle acoustic camera setup. Success across these two datasets highlights the strong generalizability of our model.

\begin{figure}[htbp]
\centering
\setlength{\tabcolsep}{1pt} 
\begin{tabular}{ccc}
\fbox{\includegraphics[width=0.3\linewidth]{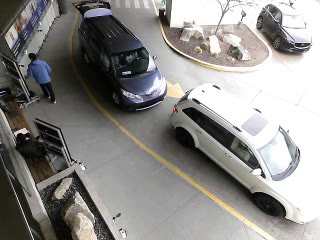}} & 
\fbox{\includegraphics[width=0.3\linewidth]{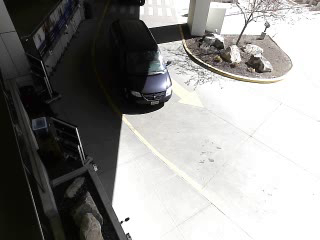}} & 
\fbox{\includegraphics[width=0.3\linewidth]{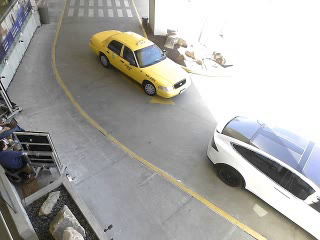}} \\ 
\fbox{\includegraphics[width=0.3\linewidth]{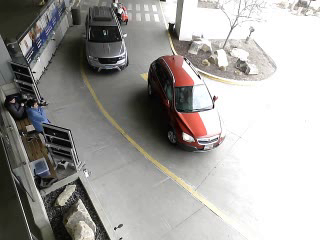}} & 
\fbox{\includegraphics[width=0.3\linewidth]{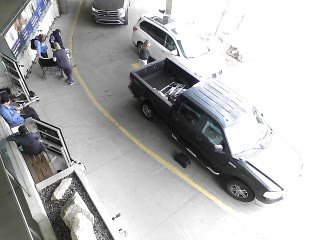}} & 
\fbox{\includegraphics[width=0.3\linewidth]{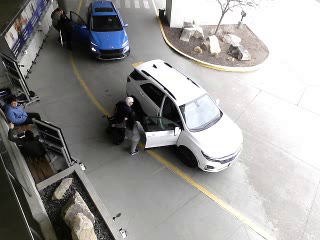}} \\
\fbox{\includegraphics[width=0.3\linewidth]{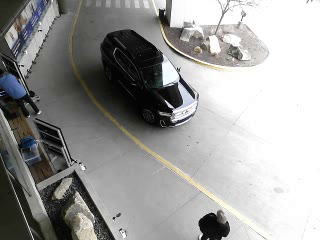}} & 
\fbox{\includegraphics[width=0.3\linewidth]{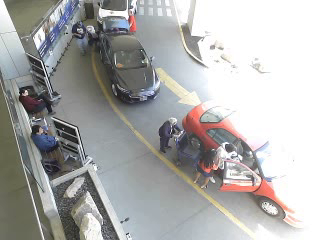}} & 
\fbox{\includegraphics[width=0.3\linewidth]{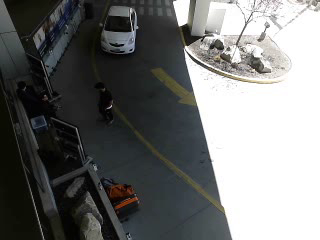}} \\
\end{tabular}
\caption{Sample images from the AVIVD dataset, illustrating vehicles of various shapes, models, colors, and sizes. The dataset also features diverse lighting conditions.}
\label{fig:AVIVD_dataset}
\end{figure}

\textbf{Audio-Visual IVD Dataset (AVIVD).} To the best of our knowledge, there is no publicly available dataset that fits our specific setup, so we created the AVIVD dataset using the method described in \cite{Li2023RealTimeIV}. Unlike \cite{MM-Distill}, our data collection setup closely follows \cite{Li2023RealTimeIV} and \cite{Gan2019SelfSupervisedMV}. As shown in \cref{fig:setup}, we mounted the webcam 20 feet above the ground and deployed an array of evenly spaced wireless microphones along the roadside. We collected recordings over four days, with four hours of data recorded each day. Data annotation for IVD is particularly challenging, as it requires tracking the engine status and precise timestamps for each vehicle. Despite having a note-taker during data collection, we developed a custom tool to label engine status by listening to the closest audio channel and cross-referencing it with the notes. To ensure that the training and validation sets contain different vehicles, we split the first 75\% of each day's recording for training and used the remaining 25\% for validation. We also resampled the data to mitigate class imbalance. We also sample per 1 second from the raw recording. The training set consists of 76,940 video clips with synchronized audio, and the validation set contains 8,431 video clips with synchronized audio. This results in 26,924 moving bounding boxes, 36,968 idling bounding boxes, and 41,868 engine-off bounding boxes in the training set, and 2,908 moving bounding boxes, 2,669 idling bounding boxes, and 3,422 engine-off bounding boxes in the validation set.

\textbf{MAVD}. MAVD \cite{MM-Distill} is an audio-visual vehicle tracking dataset consisting of 11k images and synchronized audio samples collected using an in-vehicle mounted camera and microphone array. The dataset is designed for studying audio-visual knowledge distillation in vehicle detection tasks. Unlike our setup, which uses a roadside camera and microphone array, MAVD captures data from a car-mounted perspective. The dataset covers various lighting conditions, ranging from daytime to nighttime. Our downloaded version contains 76,633 training samples and 18,873 validation samples. Since MAVD does not provide ground truth bounding boxes, we used YOLOv7 to generate soft ground truth labels for evaluation. Additionally, as our model focuses on complementary features, we excluded pure nighttime samples. After filtering, this resulted in 40,389 training samples and 10,038 validation samples.

\begin{table*}[h!]
    \centering
    \resizebox{\textwidth}{!}{ 
    \begin{tabular}{|c!{\vrule width 2pt}c!{\vrule width 2pt}c|c|c|c|c|}
    \hline
    Method & \textbf{Input Modality} & Audio Backbone & mAP@0.5 & AP Moving@0.5 & AP Idling@0.5 & AP Engine Off@0.5 \\
    \hline
    Real-Time IVD \cite{Li2023RealTimeIV} & \textcolor{red}{V+M+L} & ResNet-50 (frozen) &   \cellcolor[RGB]{240,137,153} $\textbf{80.97}$ & $92.45$ & \cellcolor[RGB]{240,137,153} $\mathbf{68.93}$ & \cellcolor[RGB]{240,137,153} $\mathbf{81.55}$ \\
    \hline
    Feature Concatenation & \textbf{\textcolor[rgb]{0,0.6,0}{V+M}} & MobileNetV3 & $77.45$ & $93.97$ & $60.35$ & $78.02$ \\
    \hline
    Feature Concatenation & \textbf{\textcolor[rgb]{0,0.6,0}{V+M}} & ResNet-50 (frozen) & $77.35$ & $93.67$ & $66.19$ & $72.18$ \\
    \hline
    AVIVDNet & \textbf{\textcolor[rgb]{0,0.6,0}{V+M}} & MobileNetV3 & $78.89$ & $90.77$ &  \cellcolor[RGB]{0,200,0} $66.81$ & \cellcolor[RGB]{0,255,0} $79.10$ \\
    \hline
    AVIVDNet & \textbf{\textcolor[rgb]{0,0.6,0}{V+M}} & ResNet-50 (frozen) & \cellcolor[RGB]{0,255,0} $79.21$ & \cellcolor[RGB]{0,238,0} $\mathbf{93.43}$ & \cellcolor[RGB]{0,200,0} $66.74$ &  $77.47$ \\
    \hline
    \end{tabular}
    } 
    \caption{Comparison with our Real-Time IVD and feature concatenation method on AVIVD Dataset on mAP and APs at IoU 0.5. We fix video backbone as MobileNetV2 width 1.0 pretrained on the Kinectic dataset for all experiments. In column Inputs, $V$, $M$, and $L$ represent video clip, audio channels, and microphone coordinates respectively.}
    \label{tab:avivd_quan_results}
\end{table*}

\subsection{Implementation Details}
Our experiments are conducted using PyTorch, the timm library \cite{rw2019timm}, and NVIDIA A6000 and Titan RTX GPUs. We used a webcam and three sets of Rode Wireless GO II microphones to collect the data. The raw video has a spatial size of $3 \times 320 \times 240$, which is reshaped to $16 \text{ (L)} \times 224 \text{ (H)} \times 224 \text{ (W)}$ before being fed into the network. The synchronized 6-channel audio $M$ consists of a 5-second audio chunk centered around the last frame of $V$, with a sample rate of 48,000 Hz. We apply a mel-spectrogram transformation to each audio channel using a window size of 1024, a hop length of 512, and 128 mel bins, resulting in a mel-spectrogram with dimensions of $128 \times 469$. The batch size is set to 16, and the learning rate is $0.0001$. Training takes 100 epochs, lasting approximately two days. For the MAVD dataset, we follow the same procedure with one exception: we resize the 8-channel spectrograms to $8 \times 768 \times 768$, as described in \cite{MM-Distill}.

We evaluate our model's performance using mean average precision (mAP) and average precision (AP). AP measures the detection accuracy for a single class by calculating the area under the precision-recall (PR) curve. True positives, false positives, and false negatives are determined based on Intersection over Union (IoU) scores. To assess overall model performance, mAP is computed as the mean of the AP values across all classes.

\begin{figure}[t!]
\centering
\begin{tabular}{c}
\begin{subfigure}[b]{0.45\textwidth}
    \centering
    \includegraphics[width=\textwidth]{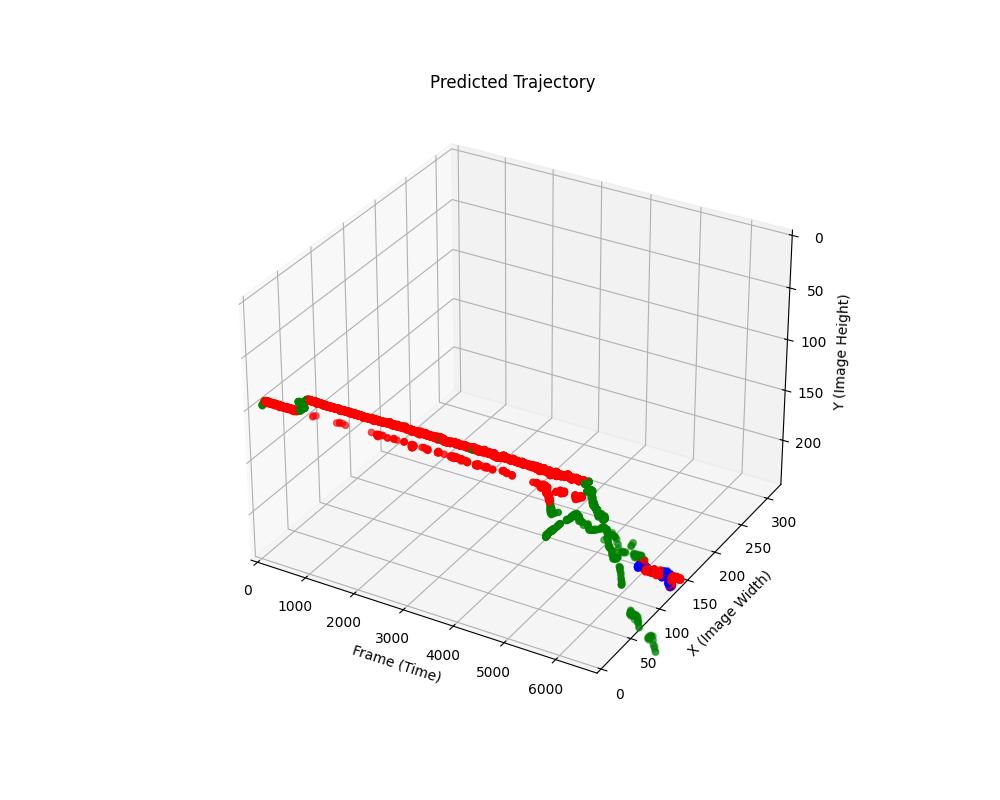}
    \caption{Predicted Trajectory}
\end{subfigure} \\
\begin{subfigure}[b]{0.45\textwidth}
    \centering
    \includegraphics[width=\textwidth]{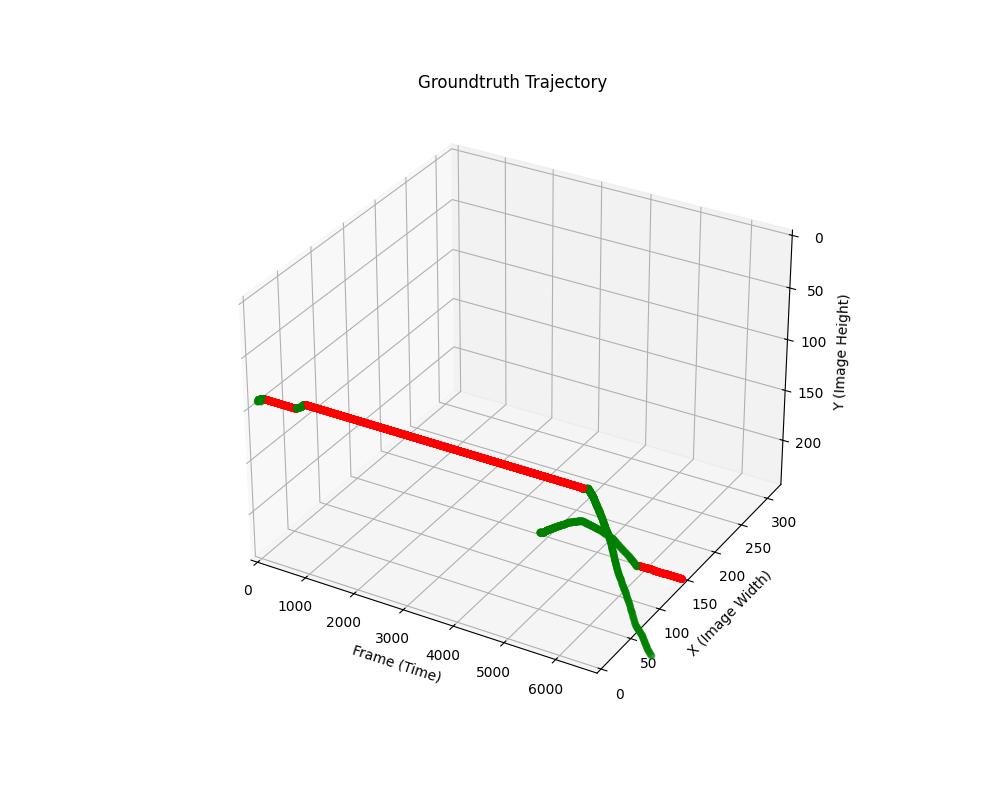}
    \caption{Ground Truth Trajectory}
\end{subfigure}
\end{tabular}
\caption{Dense Vehicle Trajectory Visualization. $X$ and $Y$ axes are aligned with image space. The rest axis is the time. Each 3D point represents the center of predicted bounding box. The color represents classes. Green is moving, red is idling, and blue is engine-off.}
\label{fig:dense_trajectory}
\end{figure}

\begin{figure*}[h!]
\centering
\begin{tabular}{@{}cc@{}}
\includegraphics[width=0.5\linewidth, height=3cm]{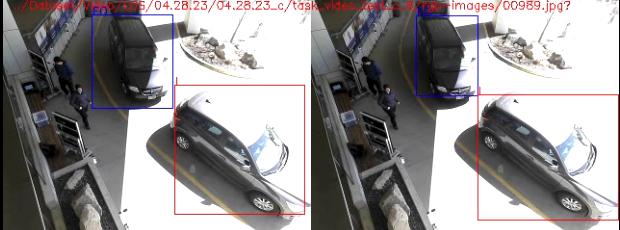} & \includegraphics[width=0.5\linewidth, height=3cm]{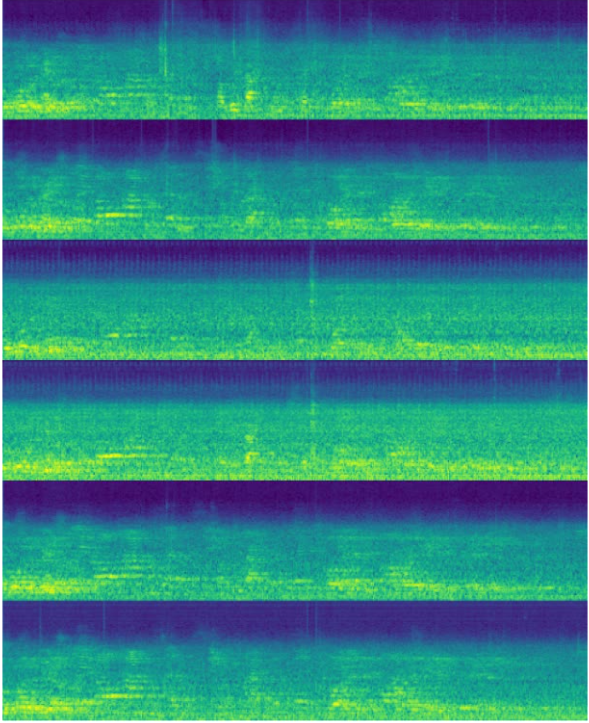} \\
\includegraphics[width=0.5\linewidth, height=3cm]{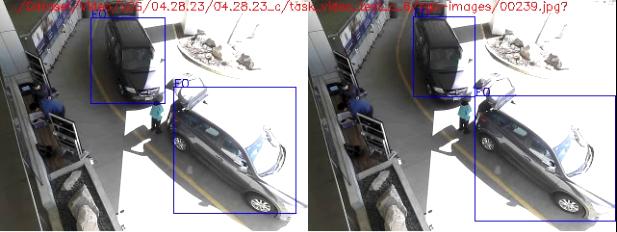} & \includegraphics[width=0.5\linewidth, height=3cm]{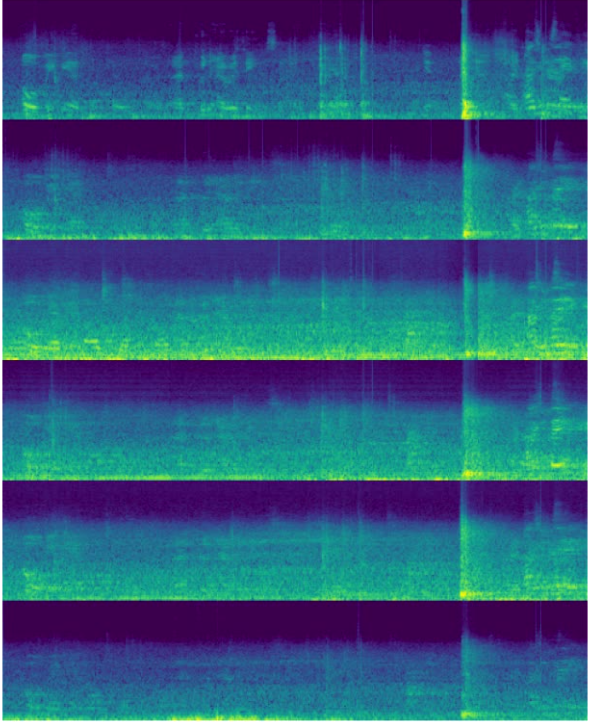} \\ 
\includegraphics[width=0.5\linewidth, height=3cm]{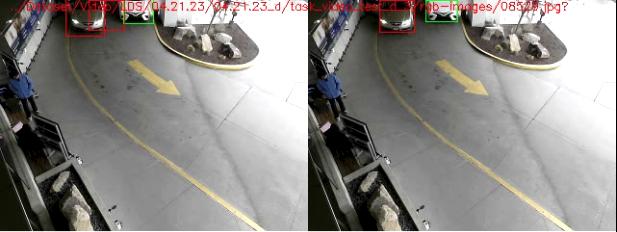} & \includegraphics[width=0.5\linewidth, height=3cm]{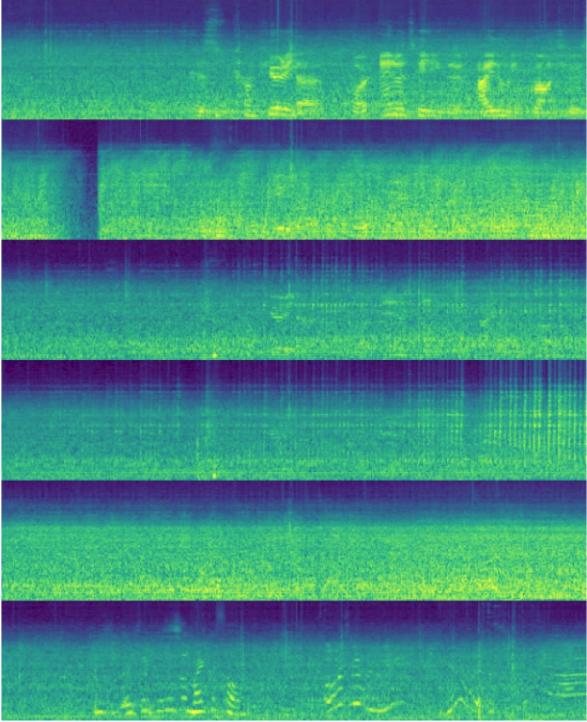} \\
\includegraphics[width=0.5\linewidth, height=3cm]{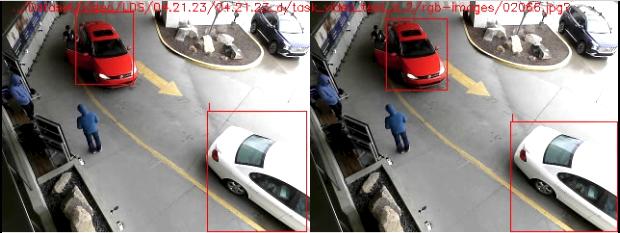} & 
\includegraphics[width=0.5\linewidth, height=3cm]{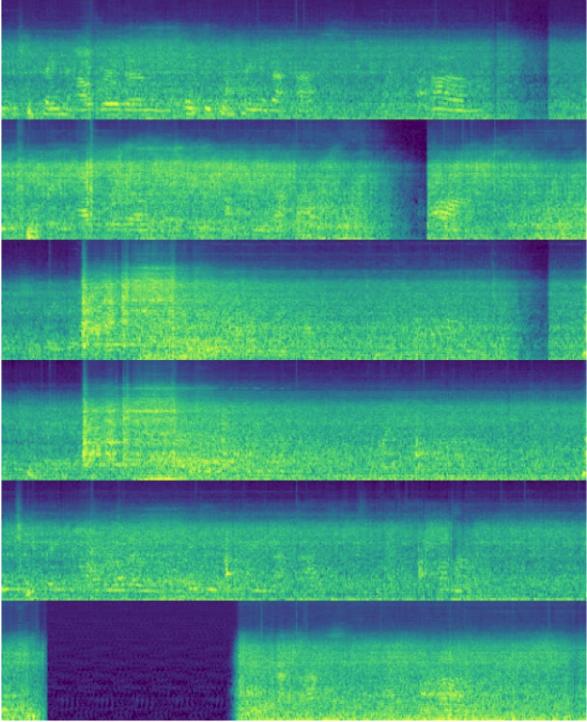} \\ 
\end{tabular}
\caption{IVD Visual Performance. The left image of each row shows the detected results (left) and the ground truth annotations (right). Green, red, and blue bounding boxes represent moving, idle, and non-idle vehicles respectively. The right image shows the corresponding spectrograms.}
\label{fig:avivd_visual_results}
\end{figure*}

\begin{figure*}[h!]
\centering
\setlength{\tabcolsep}{1pt} 
\begin{tabular}{@{}cccc@{}}
\includegraphics[width=0.25\linewidth, height=2cm]{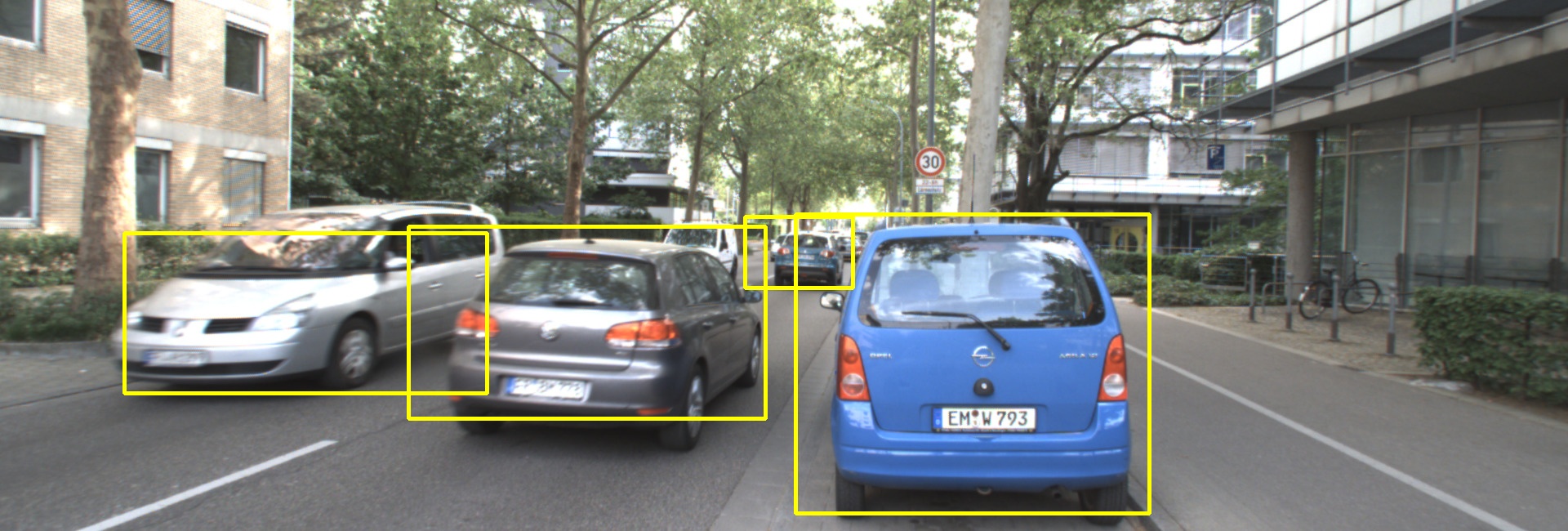} & 
\includegraphics[width=0.25\linewidth, height=2cm]{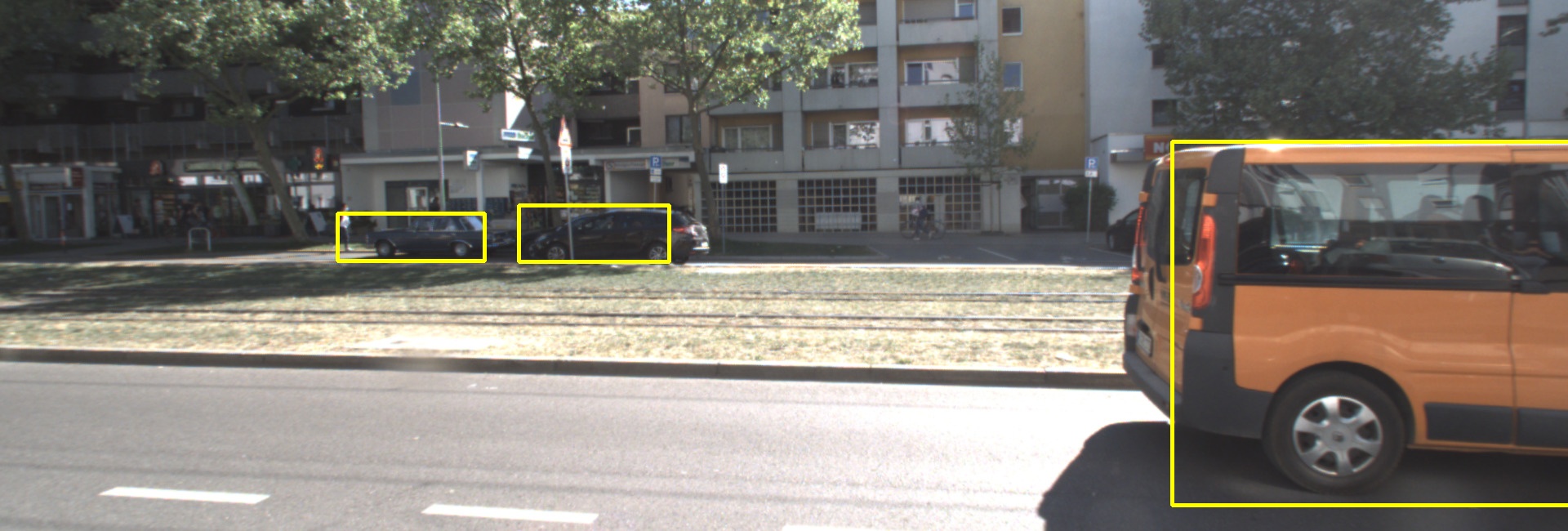} & 
\includegraphics[width=0.25\linewidth, height=2cm]{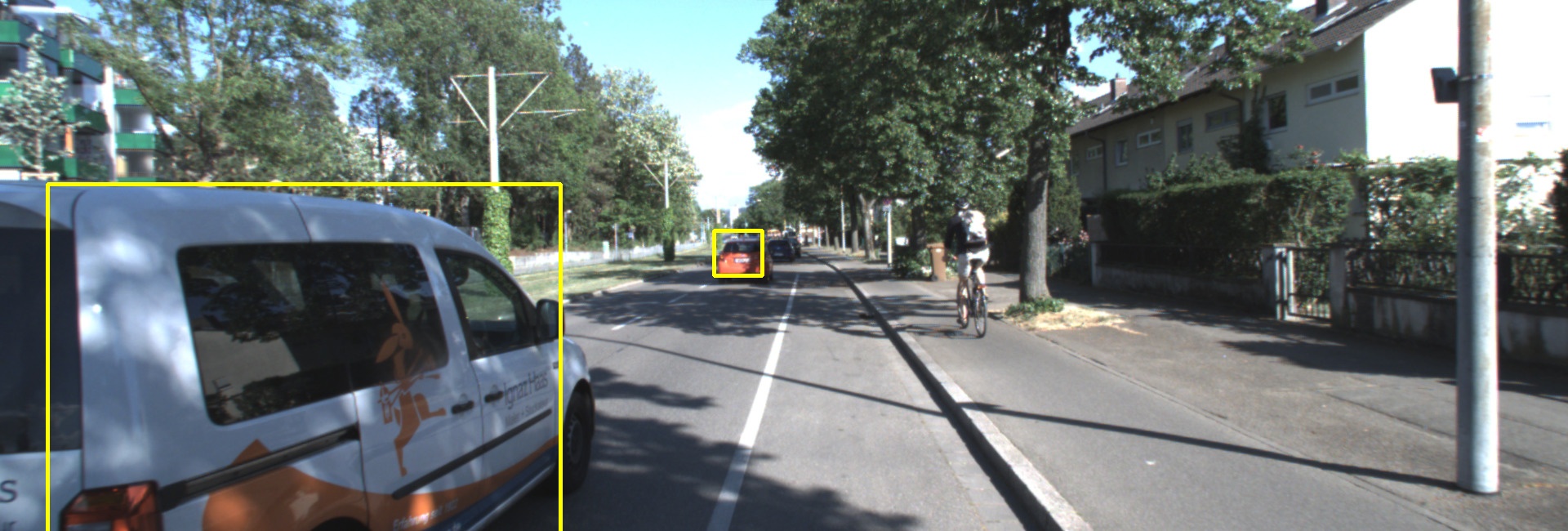} & 
\includegraphics[width=0.25\linewidth, height=2cm]{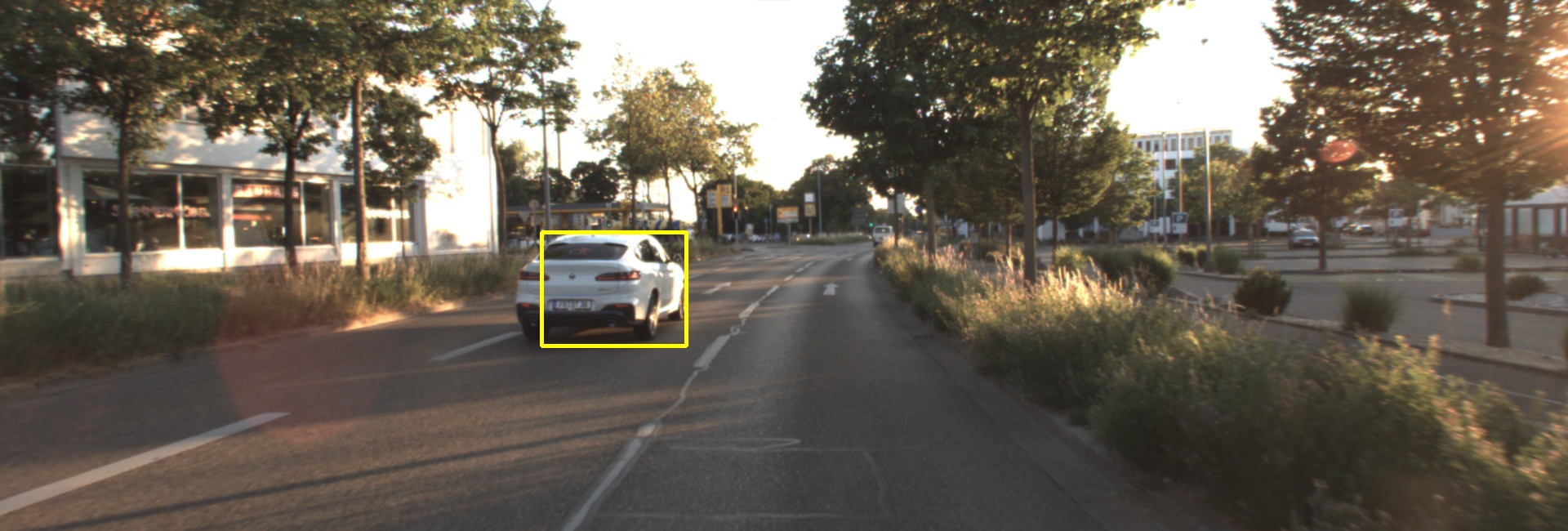} \\ 
\includegraphics[width=0.25\linewidth, height=2cm]{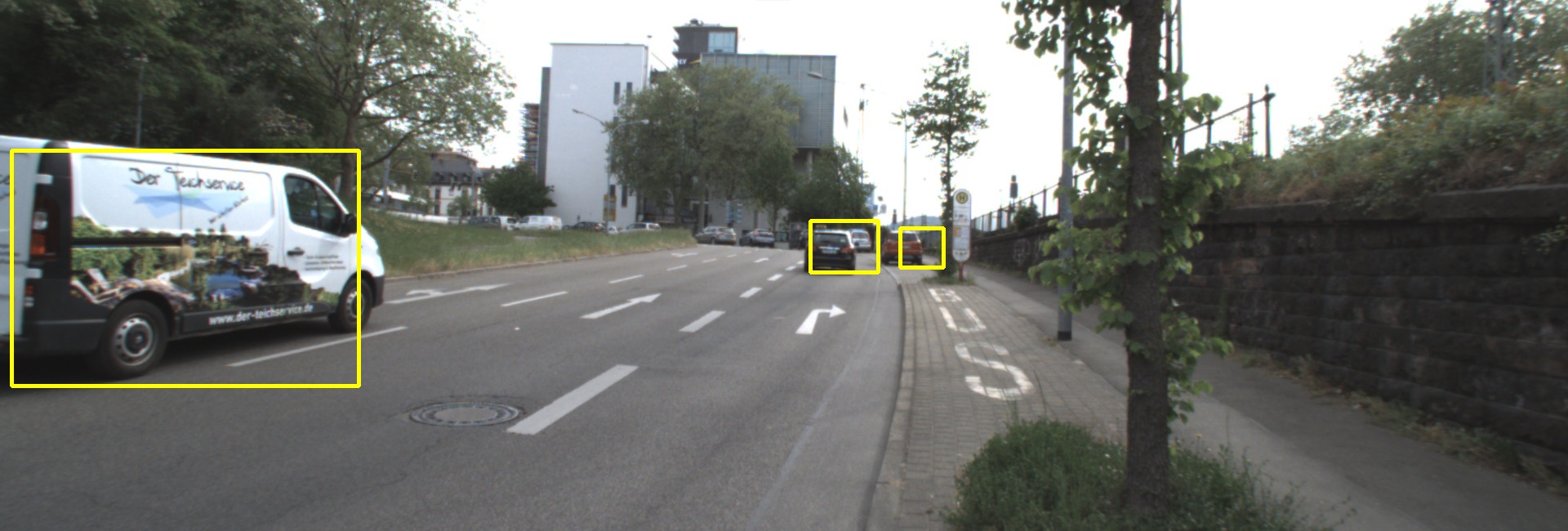} & 
\includegraphics[width=0.25\linewidth, height=2cm]{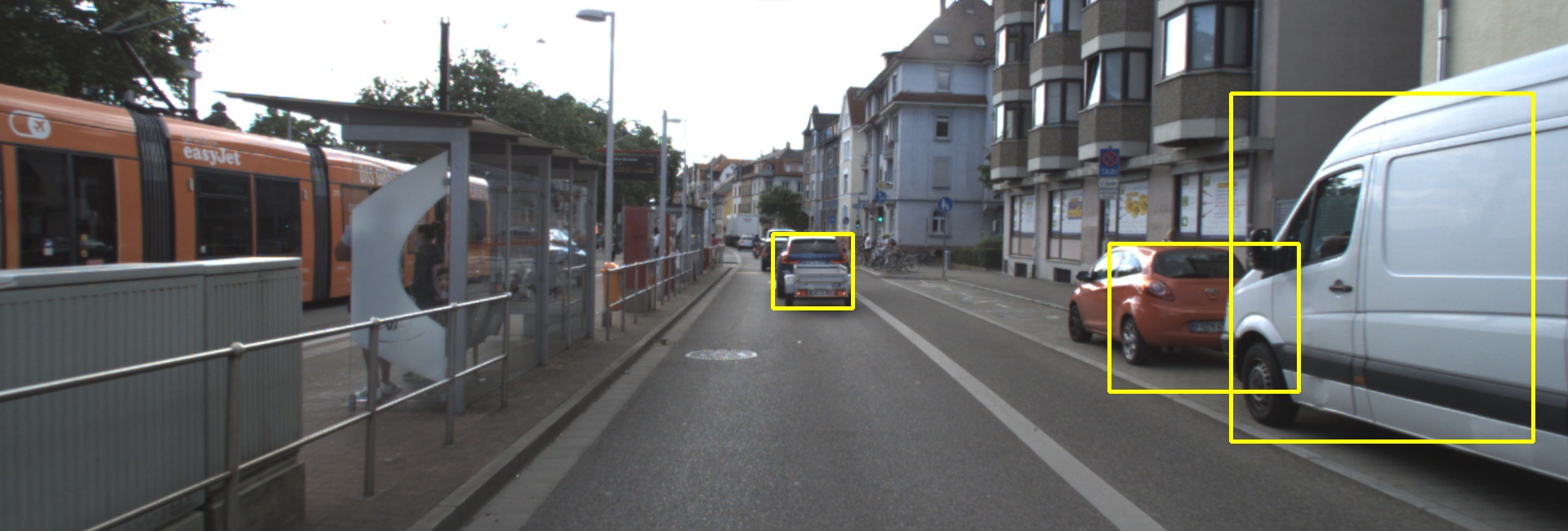} & 
\includegraphics[width=0.25\linewidth, height=2cm]{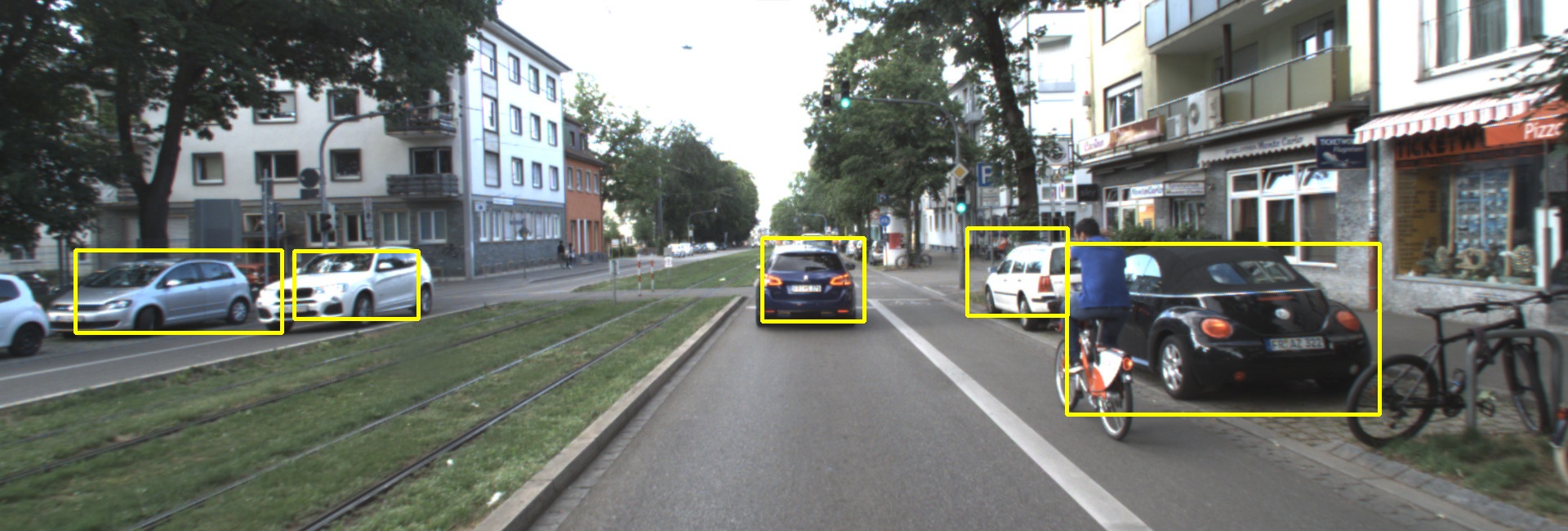} & 
\includegraphics[width=0.25\linewidth, height=2cm]{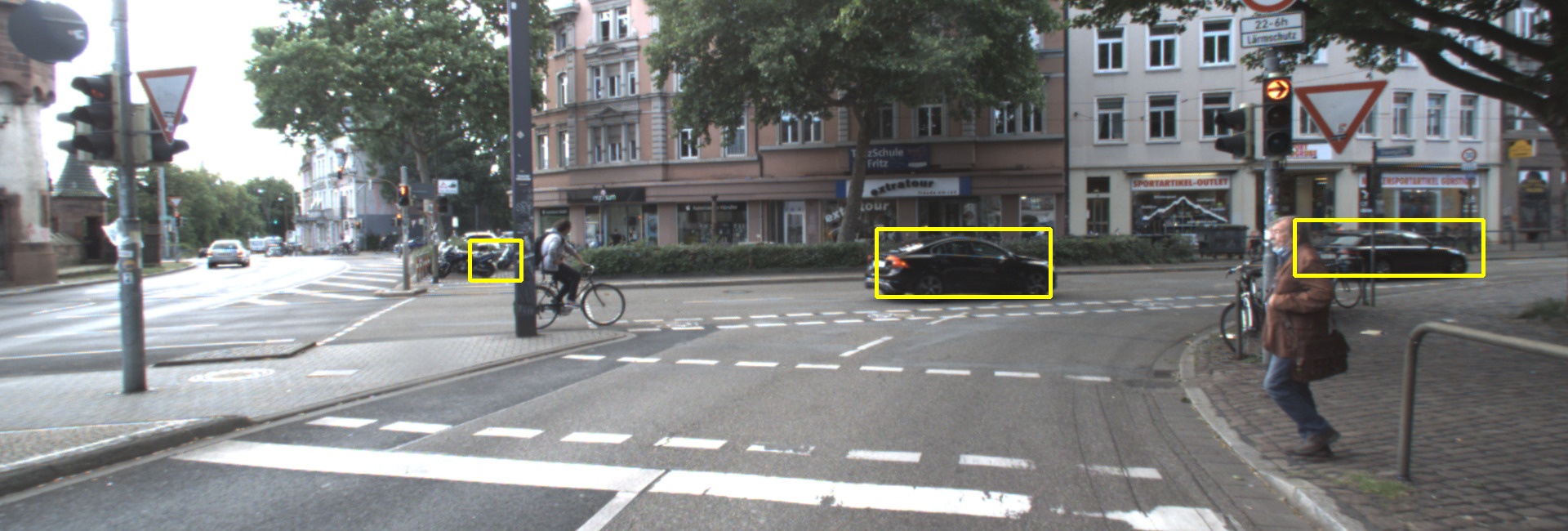} \\ 
\includegraphics[width=0.25\linewidth, height=2cm]{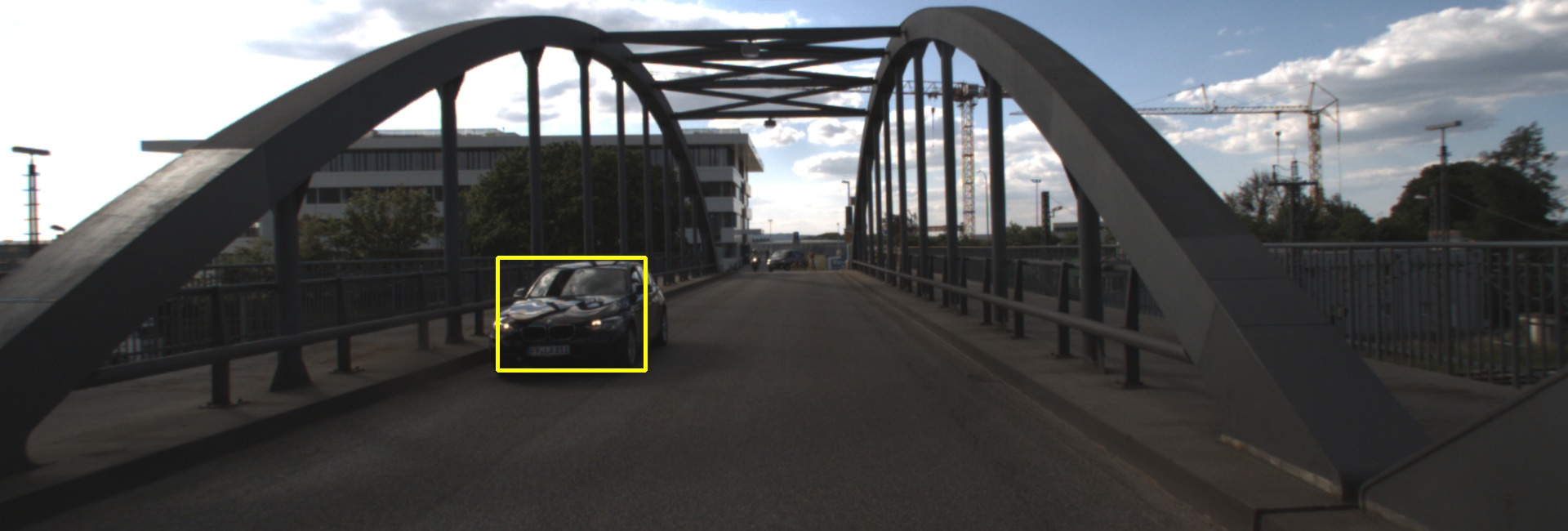} & 
\includegraphics[width=0.25\linewidth, height=2cm]{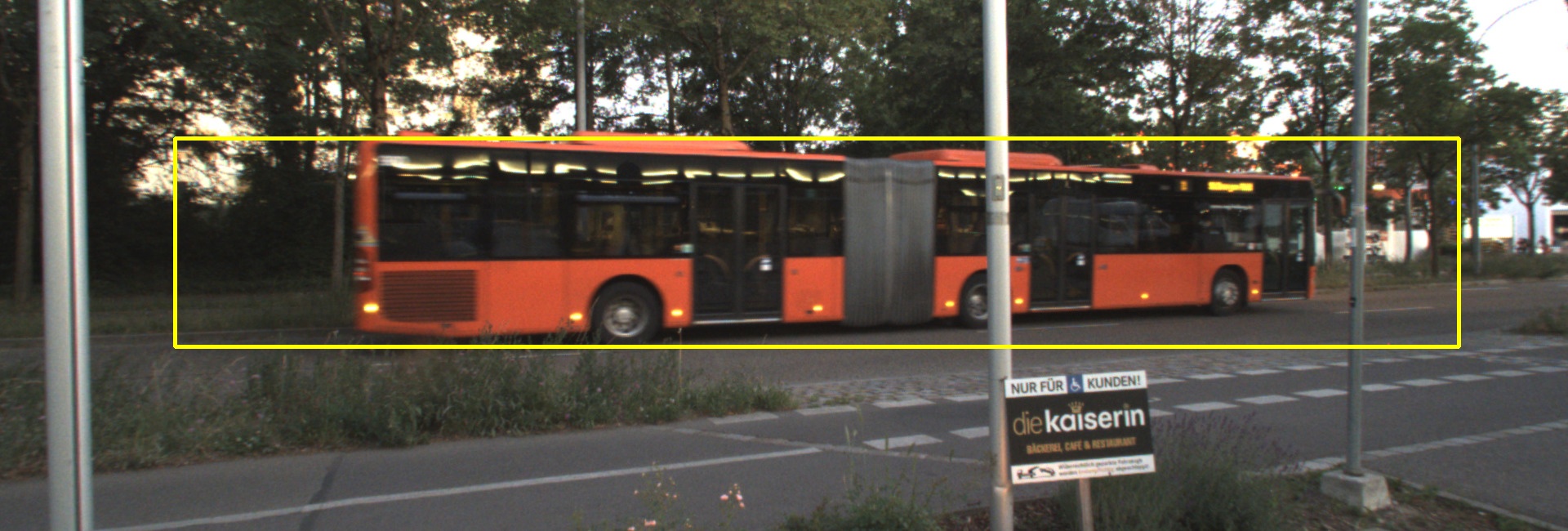} & 
\includegraphics[width=0.25\linewidth, height=2cm]{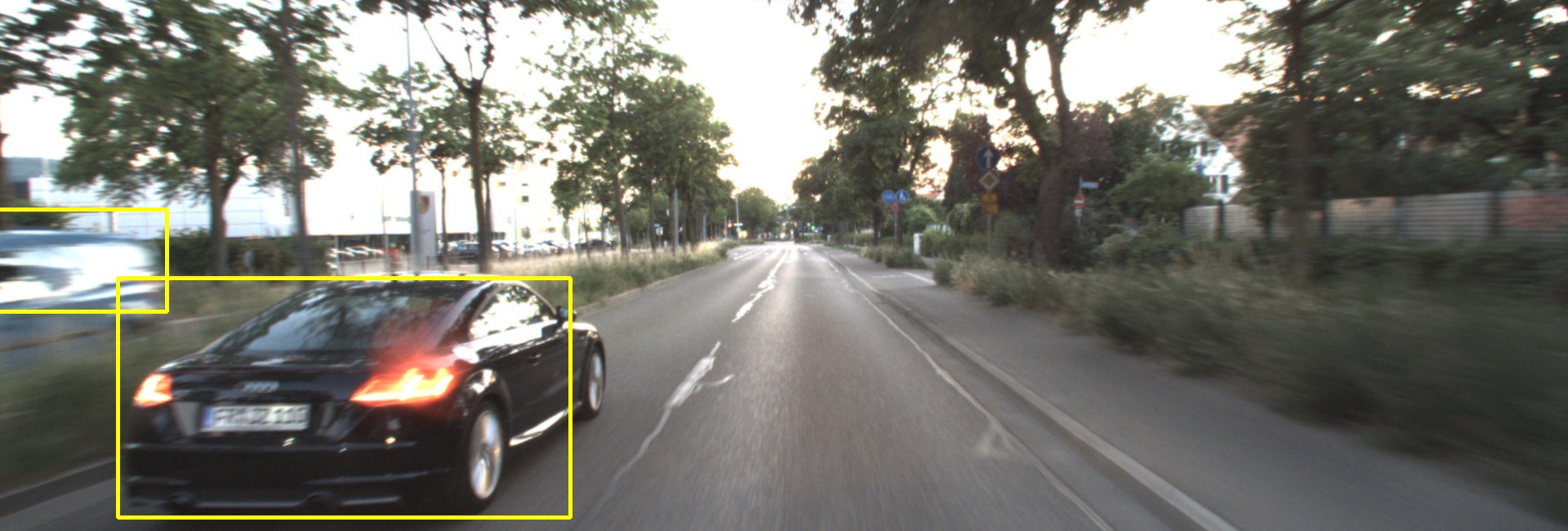} & 
\includegraphics[width=0.25\linewidth, height=2cm]{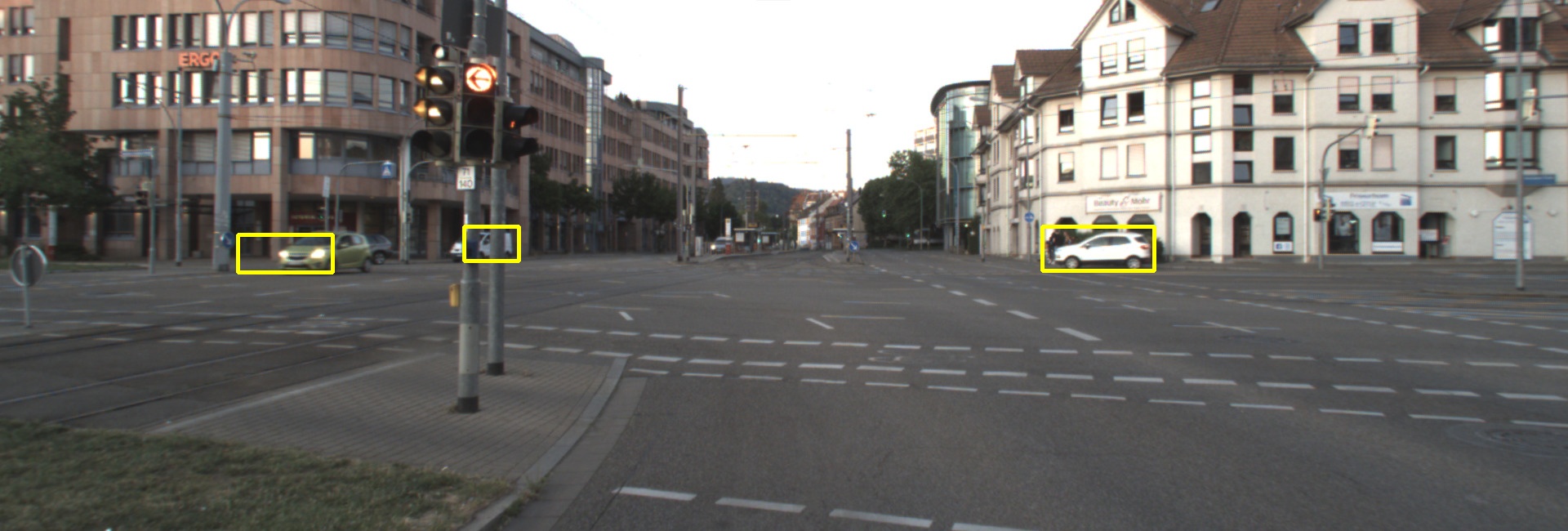} \\ 
\includegraphics[width=0.25\linewidth, height=2cm]{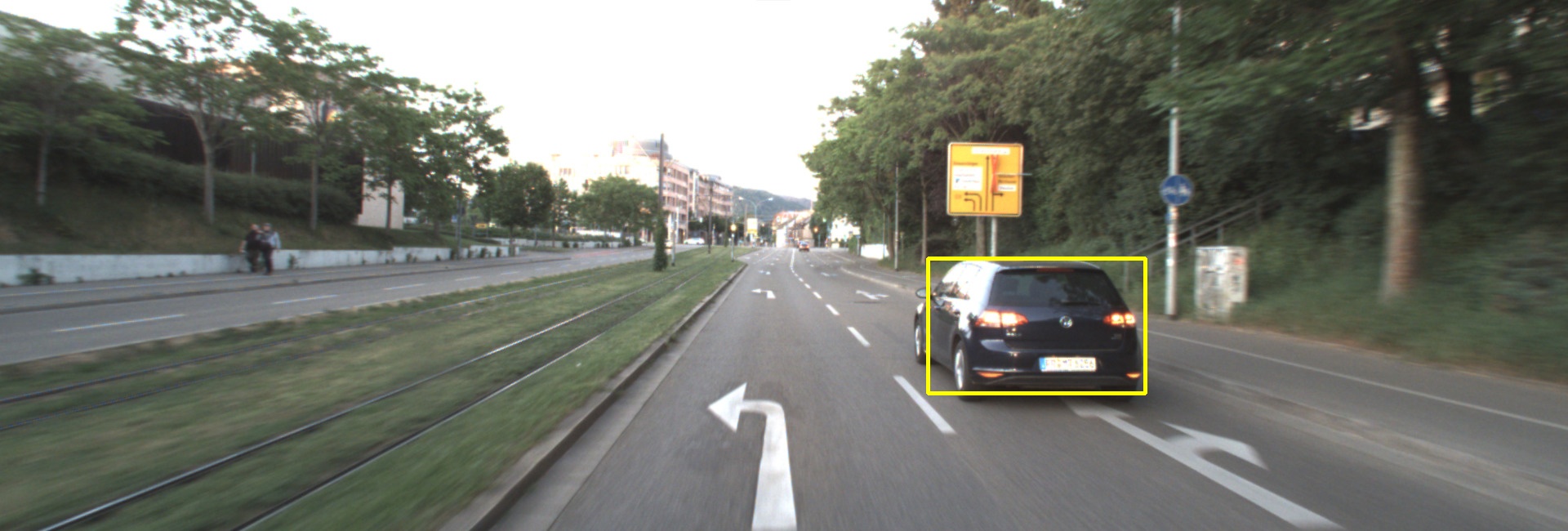} & 
\includegraphics[width=0.25\linewidth, height=2cm]{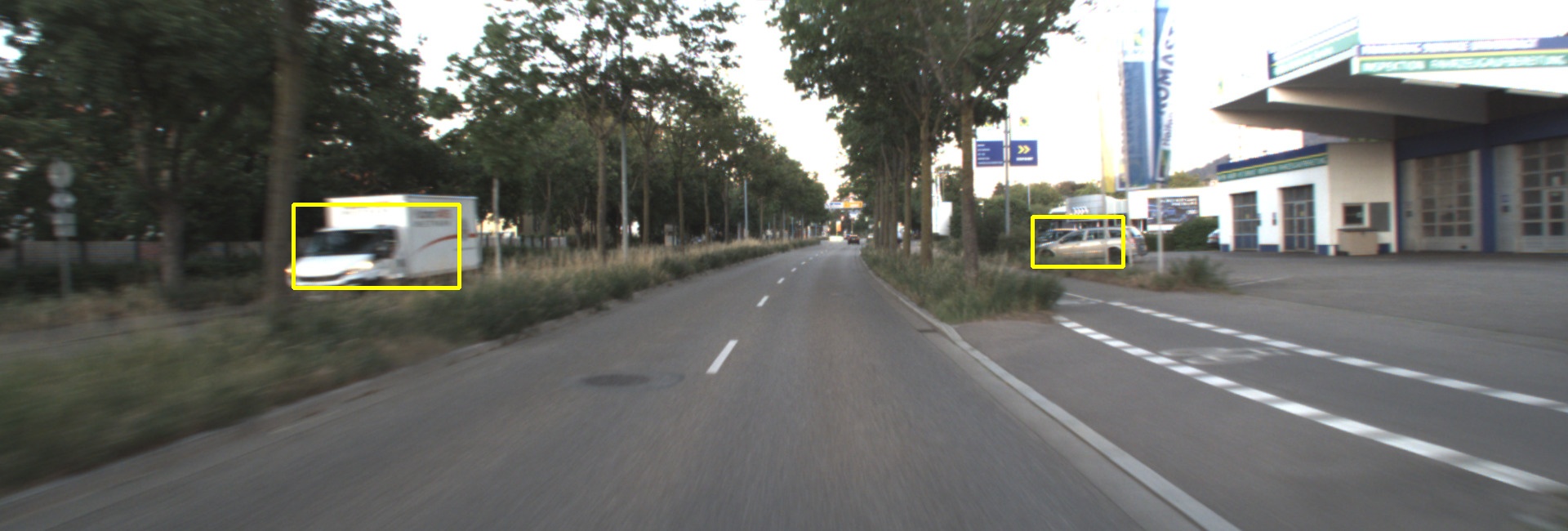} & 
\includegraphics[width=0.25\linewidth, height=2cm]{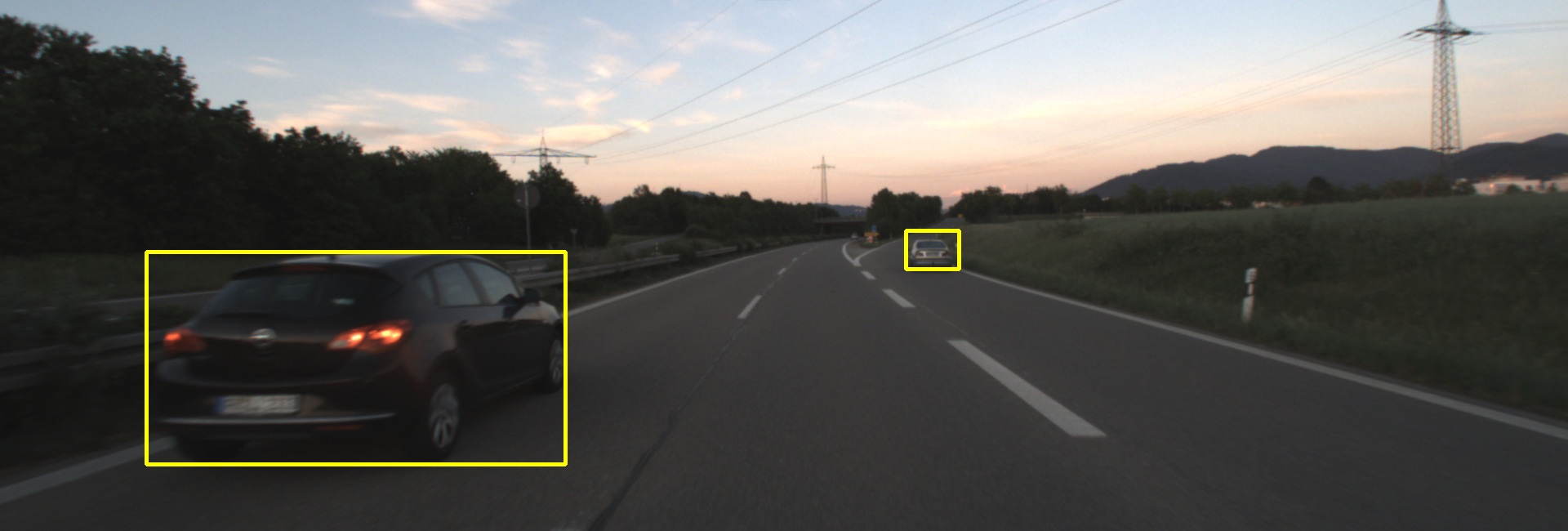} & 
\includegraphics[width=0.25\linewidth, height=2cm]{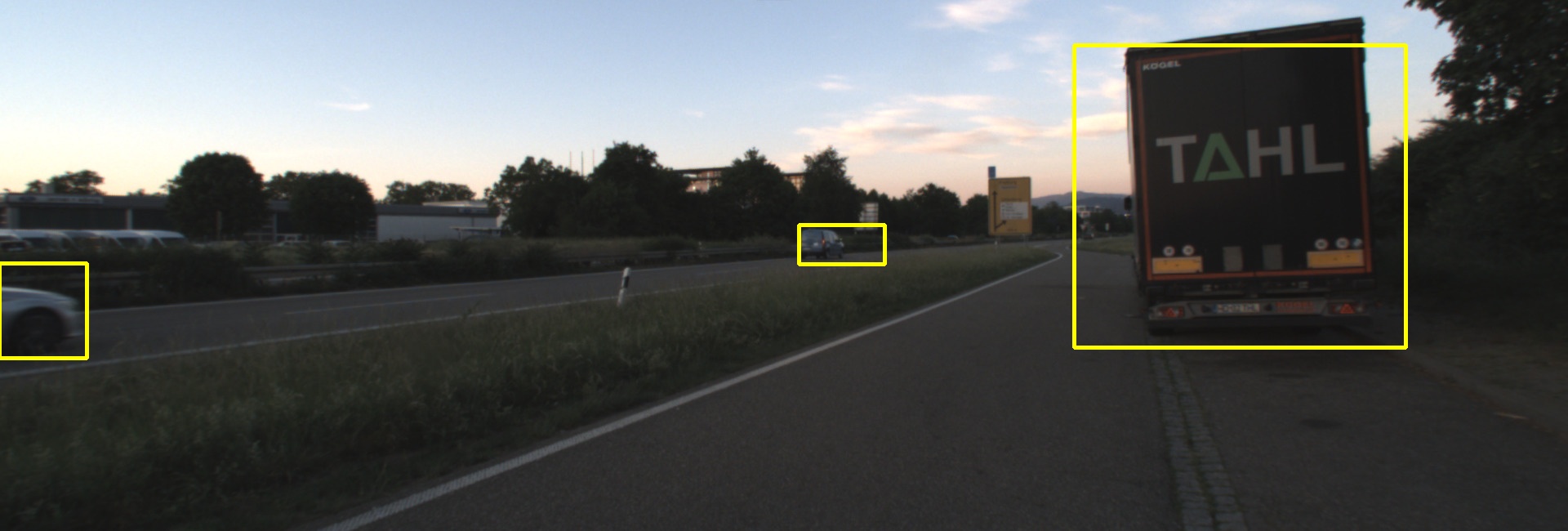} \\ 
\end{tabular}
\caption{AVIVDNet qualitative results on MAVD validation dataset. The examples illustrate detections across various roadways, times of day, traffic conditions, and vehicle types.}
\label{fig:MAVD_visual_results}
\end{figure*}

\subsection{Experiment Analysis: AVIVD}
\label{para:AVIVD_into}
We compare AVIVDNet with Real-Time IVD \cite{Li2023RealTimeIV} and feature concatenation, as shown in \cref{tab:avivd_quan_results}. AVIVDNet achieves performance comparable to Real-Time IVD while eliminating the need for user input $L$. Following \cite{Li2023RealTimeIV}, we train a ResNet-50 in a contrastive learning manner on the ESC-50 dataset \cite{piczak2015dataset} as the audio encoder and freeze it during AVIVDNet training. We can conclude four points from the table: (1)  AVIVDNet shows comparable performance to Real-Time IVD \cite{Li2023RealTimeIV} in terms of mAP. Especially when using ResNet-50 as the audio backbone, AVIVDNet achieves a mAP of 79.21\%, which is close to Real-Time IVD’s 80.97\%. (2) AVIVDNet outperforms the baseline in AP Moving, indicating excellent performance in detecting moving vehicles, particularly with the ResNet-50 backbone. This improvement is due to our joint latent space, which provides richer information for tracking vehicles. The baseline model relies solely on visual cues for motion detection. While visual data alone can be sufficient, incorporating audio information offers significant advantages. (3)  AVIVDNet shows slight improvement in detecting idling vehicles, especially with MobileNetV3, achieving 66.81\%, which is higher than feature concatenation 60.35\% and 66.19\%. This demostrates the effectiveness of our bi-directional attention module. Although it is slightly lower than Real-Time IVD (68.93\%), AVIVDNet still performs well. (4) AVIVDNet performs similarly to Real-Time IVD in AP Engine off. With MobileNetV3, it achieves 79.10\%, while Real-Time IVD reaches 81.55\%, showing strong performance in detecting vehicles with engines turned off.

We visualize our results in \cref{fig:avivd_visual_results}, which also includes 6-channel power spectrograms on the right to offer an additional perspective on engine status. When a vehicle starts its engine, the spectrogram becomes lighter, indicating increased power compared to when the engine is off. In this real-world dataset, various factors can easily cause models to fail, but AVIVDNet demonstrates its robustness in multi-vehicle scenarios. (1) Engine Switch: The first two rows illustrate an engine switch scenario, where the gray SUV in the bottom right turns off its engine. Our model accurately captures this activity, even with another vehicle nearby. (2) Side Moving Vehicle: In the third row, a gray sedan is idling while a white SUV drives past. The network experiences minor interference from the moving vehicle but still performs well. (3) Two Idling Vehicles: The fourth row shows two idling vehicles. Since our model was trained with a focus on vehicles within the driving lanes, it is not affected by the black SUV in the top-right corner, which is outside the area of interest.

\subsubsection{Dense Trajectory Visualization}
AVIVDNet also performs well in dense trajectory tracking. AVIVD dataset is sparsely sampled to facilitate training, since consecutive frames in the raw recordings do not differ significantly. However, in a real-time deployment scenario, the system would infer results on every single frame. To evaluate AVIVDNet’s ability to reconstruct vehicle trajectories, we visualized dense frames (25 FPS) in \cref{fig:dense_trajectory}. This scenario shows one vehicle entering the frame, idling, and then exiting at the bottom, while another vehicle appears near the end of the video and remains idling. By comparing the shape and color of the trajectories, we can see that AVIVDNet successfully reconstructs both vehicles' trajectories, as illustrated in \cref{fig:dense_trajectory} (a).

AVIVDNet is not without its limitations. As shown in \cref{tab:avivd_quan_results}, the AP for the idling class is lower than the baseline. We found that this shortfall primarily arises from a specific mode of failure—incorrectly detecting idling vehicle sounds. In some cases, the model misclassifies the entire idling status of two vehicles as engine off. This highlights two issues: (1) our audio encoder is still not robust enough to accurately capture all vehicle sounds in real-world conditions, and (2) using a shared feature space for both detection and classification complicates the task, leading to lower classification accuracy. Future research could explore off-the-shelf detectors for handling the detection component, allowing for improved overall performance.


\begin{table}[t]
    \centering
    \begin{tabular}{|c!{\vrule width 1pt}c|c|}
        \hline 
        Method & \begin{tabular}{c} 
        Test \\
        Modality
        \end{tabular} & mAP@0.5  \\
        \hline 
        StereoSoundNet \cite{Gan2019SelfSupervisedMV}  & $A$ & 62.38  \\
        \hline 
        Pairwise loss \cite{Liu2019StructuredKD} & $A$  & 59.72  \\
        \hline 
        AFD loss \cite{Wang2020PayAT}  & $A$  & 62.00  \\
        \hline 
        MM-DistillNet \cite{MM-Distill}  & $A$  & 62.66  \\
        \noalign{\global\arrayrulewidth=1.5pt} 
        \hline
        \noalign{\global\arrayrulewidth=0.4pt} 
        AVIVDNet & $A$+$V$ & 55.41  \\
        \hline
    \end{tabular}
    \caption{Comparison of MAVD methods on mAP metric.}
    \label{tab:MAVD_quan}
\end{table}

\subsection{Experiment Analysis: MAVD}
We also evaluated our model's detection performance on the public audio-visual vehicle tracking dataset MAVD, which has a completely different setup from ours. \cref{fig:MAVD_visual_results} illustrates predicted bounding boxes across various vehicle types, from small sedans to buses, and under diverse lighting conditions, ranging from daytime to evening. As shown in \cref{tab:MAVD_quan}, the mAP of AVIVDNet is comparable to other state-of-the-art (SOTA) models, demonstrating the robustness and effectiveness of our approach across two distinct tasks and environments. This confirms its potential for audio-visual detection and highlights its promise for integration into multi-modal self-driving car systems.

\section{Conclusion}
In this paper, we address the audio-visual complementary problem in idling vehicle detection (IVD) by proposing AVIVDNet, a novel network with a streamlined use case for fully automatic deployment. Additionally, we introduce the AVIVD dataset specifically designed for the IVD problem. AVIVDNet demonstrates comparable performance to previous methods on both the AVIVD and MAVD datasets and achieves similar results in feature co-occurrence vehicle tracking. Moreover, our method proves effective not only in surveillance camera setups but also in in-vehicle camera environments. Through this study, we observe that predicting idling and non-idling labels requires more specialized features. In particular, using a shared latent space for both bounding box and label prediction may not be necessary. Decoupling bounding box prediction from label prediction by leveraging off-the-shelf object detection models could allow the network to focus solely on label classification, presenting a promising direction for future research.

\clearpage 

\phantomsection 
\addcontentsline{toc}{section}{References} 

{\small
\bibliographystyle{ieee_fullname}
\bibliography{egbib}
}

\end{document}